\documentclass[twoside]{article}
\usepackage[accepted]{aistats2e}

\usepackage[T1]{fontenc}
\usepackage[pdftex]{hyperref}
\usepackage{times}
\usepackage{graphicx}

\usepackage{float}
\usepackage{algorithmic}
\floatstyle{boxed} \floatname{algorithm}{Algorithm}
\newfloat{algorithm}{t}{loa}[section]

\usepackage{amsmath}
\usepackage{amssymb}
\usepackage{amsthm}
\usepackage[round,colon,authoryear]{natbib}  

\newtheorem{lem}{Lemma}[section]
\newtheorem{theorem}{Theorem}[section]
\newtheorem{cor}{Corollary}[section]

 \usepackage{index}
 \usepackage{color}
 \newindex{todo}{todo}{tnd}{Todo List}

\DeclareMathOperator*{\argmin}{arg\,min}

\begin{document}

%
\runningtitle{A Reduction of Imitation Learning and Structured Prediction to No-Regret Online Learning}

%

\twocolumn[

\aistatstitle{A Reduction of Imitation Learning and Structured Prediction\\to No-Regret Online Learning}



\aistatsauthor{ St\'ephane Ross \And Geoffrey J. Gordon \And J. Andrew Bagnell }

\aistatsaddress{ Robotics Institute \\
 Carnegie Mellon University \\
 Pittsburgh, PA 15213, USA \\
 stephaneross@cmu.edu \\
\And Machine Learning Department \\
 Carnegie Mellon University \\
 Pittsburgh, PA 15213, USA \\
 ggordon@cs.cmu.edu \\
\And Robotics Institute \\
 Carnegie Mellon University \\
 Pittsburgh, PA 15213, USA \\
 dbagnell@ri.cmu.edu \\ } ]

\begin{abstract}
Sequential prediction problems such as imitation learning, where future observations depend on previous predictions (actions), violate the common i.i.d. assumptions made in statistical learning. This leads to poor performance in theory and often in practice. Some recent approaches \citep{Daume09, Ross10} provide stronger guarantees in this setting, but remain somewhat unsatisfactory as they train either non-stationary or stochastic policies and require a large number of iterations. In this paper, we propose a new iterative algorithm, which trains a stationary deterministic policy, that can be seen as a no regret algorithm in an online learning setting. We show that any such no regret algorithm, combined with additional reduction assumptions, must find a policy with good performance under the distribution of observations it induces in such sequential settings. We demonstrate that this new approach outperforms previous approaches on two challenging imitation learning problems and a benchmark sequence labeling problem.
\end{abstract}

\section{INTRODUCTION}
Sequence Prediction problems arise commonly in practice. For instance, most robotic systems must be able to predict/make a sequence of actions given a sequence of observations revealed to them over time. In complex robotic systems where standard control methods fail, we must often resort to learning a controller that can make such predictions. Imitation learning techniques, where expert demonstrations of good behavior are used to learn a controller, have proven very useful in practice and have led to state-of-the art performance in a variety of applications \citep{Schaal99, Abbeel04, Ratliff06, Silver08, Argall09, Chernova09, Ross10}. A typical approach to imitation learning is to train a classifier or regressor to predict an expert's behavior given training data of the encountered observations (input) and actions (output) performed by the expert. However since the learner's prediction affects future input observations/states during execution of the learned policy, this violate the crucial i.i.d. assumption made by most statistical learning approaches.


Ignoring this issue leads to poor performance both in theory and practice \citep{Ross10}. In particular, a classifier that makes a mistake with probability $\epsilon$ under the distribution of states/observations encountered by the expert can make as many as $T^2 \epsilon$ mistakes in expectation over $T$-steps under the distribution of states the classifier itself induces \citep{Ross10}. Intuitively this is because as soon as the learner makes a mistake, it may encounter completely different observations than those under expert demonstration, leading to a compounding of errors.


Recent approaches \citep{Ross10} can guarantee an expected number of mistakes linear (or nearly so) in the task horizon $T$ and error $\epsilon$ by training over several iterations and allowing the learner to influence the input states where expert demonstration is provided (through execution of its own controls in the system). One approach \citep{Ross10} learns a non-stationary policy by training a different policy for each time step in sequence, starting from the first step. Unfortunately this is impractical when $T$ is large or ill-defined. Another approach called SMILe \citep{Ross10}, similar to SEARN \citep{Daume09} and CPI \citep{Kakade02}, trains a stationary stochastic policy (a finite mixture of policies) by adding a new policy to the mixture at each iteration of training. However this may be unsatisfactory for practical applications as some policies in the mixture are worse than others and the learned controller may be unstable.


We propose a new meta-algorithm for imitation learning which learns a stationary deterministic policy guaranteed to perform well under its induced distribution of states (number of mistakes/costs that grows linearly in $T$ and classification cost $\epsilon$). We take a reduction-based approach \citep{Beygelzimer05} that enables reusing existing supervised learning algorithms.
Our approach is simple to implement, has no free parameters except the supervised learning algorithm sub-routine, and requires a
number of iterations that scales nearly linearly with the effective horizon of the problem. It naturally handles continuous as well as discrete
predictions. Our approach is closely related to no regret online learning algorithms \citep{CesaBianchi04, Hazan06, Kakade08} (in particular \emph{Follow-The-Leader}) but better leverages the expert in our setting. Additionally, we show that any no-regret learner can be used in a particular fashion to learn a policy that achieves similar guarantees.

We begin by establishing our notation and setting, discuss related work, and then present the \textsc{DAgger} (Dataset Aggregation) method. We analyze this approach using a no-regret and a reduction approach \citep{Beygelzimer05}. Beyond the reduction analysis, we consider the sample complexity of our approach using online-to-batch \citep{CesaBianchi04} techniques. We demonstrate \textsc{DAgger} is scalable and outperforms previous approaches in practice on two challenging imitation learning problems: 1) learning to steer a car in a 3D racing game (\emph{Super Tux Kart}) and 2) and learning to play \emph{Super Mario Bros.}, given input image features and corresponding actions by a human expert and near-optimal planner respectively. Following \citet{Daume09} in treating structured prediction as a degenerate imitation learning problem, we apply \textsc{DAgger} to the OCR \citep{Taskar03} benchmark prediction problem achieving results competitive with the state-of-the-art \citep{Taskar03, Ratliff07, Daume09} using only single-pass, greedy prediction.
\section{PRELIMINARIES}
We begin by introducing notation relevant to our setting. We denote by $\Pi$ the class of policies the learner is considering and $T$ the task horizon. For any policy $\pi$, we let $d^t_\pi$ denote the distribution of states at time $t$ if the learner executed policy $\pi$ from time step $1$ to $t-1$. Furthermore, we denote $d_\pi = \frac{1}{T} \sum_{t=1}^T d^t_\pi$ the average distribution of states if we follow policy $\pi$ for $T$ steps. Given a state $s$, we denote $C(s,a)$ the expected immediate cost of performing action $a$ in state $s$ for the task we are considering and denote $C_\pi(s) = \mathbb{E}_{a \sim \pi(s)}[C(s,a)]$ the expected immediate cost of $\pi$ in $s$. We assume $C$ is bounded in $[0,1]$. The total cost of executing policy $\pi$ for $T$-steps (\emph{i.e.}, the cost-to-go) is denoted $J(\pi) = \sum_{t=1}^T \mathbb{E}_{s \sim d^t_\pi}[C_\pi(s)] = T \mathbb{E}_{s \sim d_\pi}[C_\pi(s)]$.

In imitation learning, we may not necessarily know or observe true costs $C(s,a)$ for the particular task. Instead, we observe expert demonstrations and seek to bound $J(\pi)$ for any cost function $C$ based on how well $\pi$ mimics the expert's policy $\pi^*$. Denote $\ell$ the observed surrogate loss function we minimize instead of $C$. For instance $\ell(s,\pi)$ may be the expected 0-1 loss of $\pi$ with respect to $\pi^*$ in state $s$, or a squared/hinge loss of $\pi$ with respect to $\pi^*$ in $s$. Importantly, in many instances, $C$ and $\ell$ may be the same function-- for instance, if we are interested in optimizing the learner's ability to predict the actions chosen by an expert.

Our goal is to find a policy $\hat{\pi}$ which minimizes the observed surrogate loss under its induced distribution of states, i.e.:
\begin{equation}
\hat{\pi} = \argmin_{\pi \in \Pi} \mathbb{E}_{s \sim d_\pi}[\ell(s,\pi)]
\end{equation}
As system dynamics are assumed both unknown and complex, we cannot compute $d_\pi$ and can only sample it by executing $\pi$ in the system. Hence this is a non-i.i.d. supervised learning problem due to the dependence of the input distribution on the policy $\pi$ itself. The interaction between
policy and the resulting distribution makes optimization difficult as it results in a non-convex objective even if the loss $\ell(s,\cdot{})$ is convex in $\pi$ for all states $s$. We now briefly review previous approaches and their guarantees.
\subsection{Supervised Approach to Imitation}
The traditional approach to imitation learning ignores the change in distribution and simply trains a policy $\pi$ that performs well under the distribution of states encountered by the expert $d_{\pi^*}$. This can be achieved using any standard supervised learning algorithm. It finds the policy $\hat{\pi}_{sup}$:
\begin{equation}
\hat{\pi}_{sup} = \argmin_{\pi \in \Pi} \mathbb{E}_{s \sim d_{\pi^*}}[\ell(s,\pi)]
\end{equation}
Assuming $\ell(s,\pi)$ is the 0-1 loss (or upper bound on the 0-1 loss) implies the following performance guarantee with respect to any task cost function $C$ bounded in $[0,1]$:
\begin{theorem}
\citep{Ross10} Let $\mathbb{E}_{s \sim d_{\pi^*}}[\ell(s,\pi)] = \epsilon$, then $J(\pi) \leq J(\pi^*) + T^2 \epsilon$.
\begin{proof}
Follows from result in \citet{Ross10} since $\epsilon$ is an upper bound on the 0-1 loss of $\pi$ in $d_{\pi^*}$.
\end{proof}
\end{theorem}
Note that this bound is tight, i.e. there exist problems such that a policy $\pi$ with $\epsilon$ 0-1 loss on $d_{\pi^*}$ can incur extra cost that grows quadratically in $T$. \citet{Kaariainen06} demonstrated this in a sequence prediction setting\footnote{In their example, an error rate of $\epsilon > 0$ when trained to predict the next output in sequence with the previous correct output as input can lead to an expected number of mistakes of $\frac{T}{2} - \frac{1-(1-2\epsilon)^{T+1}}{4\epsilon} + \frac{1}{2}$ over sequences of length $T$ at test time. This is bounded by $T^2 \epsilon$ and behaves as $\Theta(T^2\epsilon)$ for small $\epsilon$.} and \citet{Ross10} provided an imitation learning example where $J(\hat{\pi}_{sup}) = (1-\epsilon T)J(\pi^*) + T^2 \epsilon$. Hence the traditional supervised learning approach has poor performance guarantees due to the quadratic growth in $T$. Instead we would prefer approaches that can guarantee growth linear or near-linear in $T$ and $\epsilon$. The following two approaches from \citet{Ross10} achieve this on some classes of imitation learning problems, including all those where surrogate loss $\ell$ upper bounds $C$.
%
%
\subsection{Forward Training}
The forward training algorithm introduced by \citet{Ross10} trains a non-stationary policy (one policy $\pi_t$ for each time step $t$) iteratively over $T$ iterations, where at iteration $t$, $\pi_t$ is trained to mimic $\pi^*$ on the distribution of states at time $t$ induced by the previously trained policies $\pi_1,\pi_2,\dots,\pi_{t-1}$. By doing so, $\pi_t$ is trained on the actual distribution of states it will encounter during execution of the learned policy. Hence the forward algorithm guarantees that the expected loss under the distribution of states induced by the learned policy matches the average loss during training, and hence improves performance. 

We here provide a theorem slightly more general than the one provided by \citet{Ross10} that applies to any policy $\pi$ that can guarantee $\epsilon$ surrogate loss under its own distribution of states. This will be useful to bound the performance of our new approach presented in Section \ref{secDAgger}.

Let $Q^{\pi'}_t(s,\pi)$ denote the $t$-step cost of executing $\pi$ in initial state $s$ and then following policy $\pi'$ and assume $\ell(s,\pi)$ is the 0-1 loss (or an upper bound on the 0-1 loss), then we have the following performance guarantee with respect to any task cost function $C$ bounded in $[0,1]$:
\begin{theorem} \label{thmforward}
Let $\pi$ be such that $\mathbb{E}_{s \sim d_{\pi}}[\ell(s,\pi)] = \epsilon$, and $Q^{\pi^*}_{T-t+1}(s,a) - Q^{\pi^*}_{T-t+1}(s,\pi^*) \leq u$ for all action $a$, $t \in \{1,2,\dots,T\}$, $d^{t}_\pi(s) > 0$, then $J(\pi) \leq J(\pi^*) + u T \epsilon$.
\begin{proof}
We here follow a similar proof to \citet{Ross10}. Given our policy $\pi$, consider the policy $\pi_{1:t}$, which executes $\pi$ in the first $t$-steps and then execute the expert $\pi^*$. Then
\begin{displaymath}
\begin{array}{rl}
\multicolumn{2}{l}{J(\pi)}\\
= & J(\pi^*) + \sum_{t=0}^{T-1} [J(\pi_{1:T-t}) - J(\pi_{1:T-t-1})]\\
= & J(\pi^*) + \sum_{t=1}^{T} \mathbb{E}_{s \sim d^t_\pi}[Q^{\pi^*}_{T-t+1}(s,\pi) - Q^{\pi^*}_{T-t+1}(s,\pi^*)]\\
\leq & J(\pi^*) + u \sum_{t=1}^{T} \mathbb{E}_{s \sim d^t_\pi}[\ell(s,\pi)]\\
= & J(\pi^*) + u T \epsilon\\
\end{array}
\end{displaymath}
The inequality follows from the fact that $\ell(s,\pi)$ upper bounds the 0-1 loss, and hence the probability $\pi$ and $\pi^*$ pick different actions in $s$; when they pick different actions, the increase in cost-to-go $\leq u$.
\end{proof}
\end{theorem}
%
%
In the worst case, $u$ could be $O(T)$ and the forward algorithm wouldn't provide any improvement over the traditional supervised learning approach. However, in many cases $u$ is $O(1)$ or sub-linear in $T$ and the forward algorithm leads to improved performance. For instance if $C$ is the 0-1 loss with respect to the expert, then $u \leq 1$. Additionally if $\pi^*$ is able to recover from mistakes made by $\pi$, in the sense that within a few steps, $\pi^*$ is back in a distribution of states that is close to what $\pi^*$ would be in if $\pi^*$ had been executed initially instead of $\pi$, then $u$ will be $O(1)$. \footnote{This is the case for instance in Markov Desision Processes (MDPs) when the Markov Chain defined by the system dynamics and policy $\pi^*$ is rapidly mixing. In particular, if it is $\alpha$-mixing with exponential decay rate $\delta$ then $u$ is $O(\frac{1}{1-\exp(-\delta)})$.} A drawback of the forward algorithm is that it is impractical when $T$ is large (or undefined) as we must train $T$ different policies sequentially and cannot stop the algorithm before we complete all $T$ iterations. Hence it can not be applied to most real-world applications.
\subsection{Stochastic Mixing Iterative Learning}
SMILe, proposed by \citet{Ross10},  alleviates this problem and can be applied in practice when $T$ is large or undefined by adopting an approach similar to SEARN \citep{Daume09} where a stochastic stationary policy is trained over several iterations. Initially SMILe starts with a policy $\pi_0$ which always queries and executes the expert's action choice. At iteration $n$, a policy $\hat{\pi}_n$ is trained to mimic the expert under the distribution of trajectories $\pi_{n-1}$ induces and then updates $\pi_{n} = \pi_{n-1} + \alpha(1-\alpha)^{n-1}(\hat{\pi}_n - \pi_0)$. This update is interpreted as adding probability $\alpha(1-\alpha)^{n-1}$ to executing policy $\hat{\pi}_n$ at any step and removing probability $\alpha(1-\alpha)^{n-1}$ of executing the queried expert's action. At iteration $n$, $\pi_n$ is a mixture of $n$ policies and the probability of using the queried expert's action is $(1-\alpha)^{n}$. We can stop the algorithm at any iteration $N$ by returning the re-normalized policy $\tilde{\pi}_N = \frac{\pi_N - (1-\alpha)^N \pi_0}{1-(1-\alpha)^N}$ which doesn't query the expert anymore. \citet{Ross10} showed that choosing $\alpha$ in $O(\frac{1}{T^2})$ and $N$ in $O(T^2 \log T)$ guarantees near-linear regret in $T$ and $\epsilon$ for some class of problems. 
\section{DATASET AGGREGATION} \label{secDAgger}
We now present \textsc{DAgger} (Dataset Aggregation), an iterative algorithm that trains a deterministic policy that achieves good performance guarantees under its induced distribution of states.

In its simplest form, the algorithm proceeds as follows. At the first iteration, it uses the expert's policy to gather a dataset of trajectories $\mathcal{D}$ and train a policy $\hat{\pi}_2$ that best mimics the expert on those trajectories. Then at iteration $n$, it uses $\hat{\pi}_{n}$ to collect more trajectories and adds those trajectories to the dataset $\mathcal{D}$. The next policy $\hat{\pi}_{n+1}$ is the policy that best mimics the expert on the whole dataset $\mathcal{D}$. In other words, \textsc{DAgger} proceeds by collecting a dataset at each iteration under the current policy and trains the next policy under the aggregate of all collected datasets. The intuition behind this algorithm is that over the iterations, we are building up the set of inputs that the learned policy is likely to encounter during its execution based on previous experience (training iterations). This algorithm can be interpreted as a \emph{Follow-The-Leader} algorithm in that at iteration $n$ we pick the best policy $\hat{\pi}_{n+1}$ in hindsight, i.e. under all trajectories seen so far over the iterations.

To better leverage the presence of the expert in our imitation learning setting, we optionally allow the algorithm to use a modified policy  $\pi_i = \beta_i \pi^* + (1-\beta_i) \hat{\pi}_i$ at iteration $i$ that queries the expert to choose controls a fraction of the time while collecting the next dataset.
This is often desirable in practice as the first few policies, with relatively few datapoints, may make many more mistakes and visit states that are irrelevant as the policy improves.

We will typically use $\beta_1 = 1$ so that we do not have to specify an initial policy $\hat{\pi}_1$ before getting data from the expert's behavior. Then we could choose $\beta_i = p^{i-1}$ to have a probability of using the expert that decays exponentially as in SMILe and SEARN. We show below the only requirement is that $\{ \beta_i \}$ be a sequence such that $\overline{\beta}_N = \frac{1}{N}\sum_{i=1}^N \beta_i \rightarrow 0$ as $N \rightarrow \infty$. The simple, parameter-free version of the algorithm described above is the special case $\beta_i = I(i=1)$ for $I$ the indicator function, which often performs best in practice (see Section \ref{sec:exp}). The general \textsc{DAgger} algorithm is detailed in Algorithm~\ref{algDAgger}.
\begin{algorithm}
\begin{algorithmic}
\STATE Initialize $\mathcal{D} \leftarrow \emptyset$.
\STATE Initialize $\hat{\pi}_1$ to any policy in $\Pi$.
\FOR{$i=1$ \textbf{to} $N$}
\STATE Let $\pi_i = \beta_i \pi^* + (1-\beta_i) \hat{\pi}_{i}$.
\STATE Sample $T$-step trajectories using $\pi_i$.
\STATE Get dataset $\mathcal{D}_i = \{ (s, \pi^*(s)) \}$ of visited states by $\pi_i$ and actions given by expert.
\STATE Aggregate datasets: $\mathcal{D} \leftarrow \mathcal{D} \bigcup \mathcal{D}_i$.
\STATE Train classifier $\hat{\pi}_{i+1}$ on $\mathcal{D}$.
\ENDFOR
\STATE \textbf{Return} best $\hat{\pi}_i$ on validation.
\end{algorithmic}
\caption{\textsc{DAgger} Algorithm. \label{algDAgger}}
\end{algorithm}
The main result of our analysis in the next section is the following guarantee for \textsc{DAgger}. Let $\pi_{1:N}$ denote the sequence of policies $\pi_1, \pi_2, \dots, \pi_N$. Assume $\ell$ is strongly convex and bounded over $\Pi$. Suppose $\beta_i \leq (1-\alpha)^{i-1}$ for all $i$ for some constant $\alpha$ independent of $T$. Let $\epsilon_N = \min_{\pi \in \Pi} \frac{1}{N} \sum_{i=1}^N \mathbb{E}_{s \sim d_{\pi_i}}[ \ell(s,\pi) ]$ be the true loss of the best policy in hindsight. Then the following holds in the infinite sample case (infinite number of sample trajectories at each iteration):
\begin{theorem}\label{thmDagger}
For \textsc{DAgger}, if $N$ is $\tilde{O}(T)$ there exists a policy $\hat{\pi} \in \hat{\pi}_{1:N}$ s.t. $\mathbb{E}_{s \sim d_{\hat{\pi}}}[ \ell(s,\hat{\pi}) ] \leq \epsilon_N + O(1/T)$
\end{theorem}
In particular, this holds for the policy $\hat{\pi} = \argmin_{\pi \in \hat{\pi}_{1:N}} \mathbb{E}_{s \sim d_{\pi}}[ \ell(s,\pi) ]$. \footnote{It is not necessary to find the best policy in the sequence that minimizes the loss under its distribution; the same guarantee holds for the policy which uniformly randomly picks one policy in the sequence $\hat{\pi}_{1:N}$ and executes that policy for $T$ steps.}
If the task cost function $C$ corresponds to (or is upper bounded by) the surrogate loss $\ell$ then this bound tells us directly that $J(\hat{\pi}) \leq T \epsilon_N + O(1)$. For arbitrary task cost function $C$, then if $\ell$ is an upper bound on the 0-1 loss with respect to $\pi^*$, combining this result with Theorem \ref{thmforward} yields that:
\begin{theorem}\label{thmDaggerCost}
For \textsc{DAgger}, if $N$ is $\tilde{O}(uT)$ there exists a policy $\hat{\pi} \in \hat{\pi}_{1:N}$ s.t. $J(\hat{\pi}) \leq J(\pi^*) + uT\epsilon_N + O(1)$.
\end{theorem}

\paragraph{Finite Sample Results}
In the finite sample case, suppose we sample $m$ trajectories with $\pi_i$ at each iteration $i$, and denote this dataset $D_i$. Let $\hat{\epsilon}_N = \min_{\pi \in \Pi} \frac{1}{N} \sum_{i=1}^N \mathbb{E}_{s \sim D_i}[ \ell(s,\pi) ]$ be the training loss of the best policy on the sampled trajectories, then using Azuma-Hoeffding's inequality leads to the following guarantee:
\begin{theorem}\label{thmDaggerGen}
For \textsc{DAgger}, if $N$ is $O(T^2\log(1/\delta))$ and $m$ is $O(1)$ then with probability at least $1-\delta$ there exists a policy $\hat{\pi} \in \hat{\pi}_{1:N}$ s.t. $\mathbb{E}_{s \sim d_{\hat{\pi}}}[ \ell(s,\hat{\pi}) ] \leq \hat{\epsilon}_N + O(1/T)$
\end{theorem}
A more refined analysis taking advantage of the strong convexity of the loss function \citep{Kakade09} may lead to tighter generalization bounds that require $N$ only of order $\tilde{O}(T\log(1/\delta))$. Similarly:
%
\begin{theorem}\label{thmDaggerGenCost}
For \textsc{DAgger}, if $N$ is $O(u^2T^2\log(1/\delta))$ and $m$ is $O(1)$ then with probability at least $1-\delta$ there exists a policy $\hat{\pi} \in \hat{\pi}_{1:N}$ s.t. $J(\hat{\pi}) \leq J(\pi^*) + uT\hat{\epsilon}_N + O(1)$.
\end{theorem}
%
%
\section{THEORETICAL ANALYSIS}
The theoretical analysis of \textsc{DAgger} only relies on the no-regret property of the underlying \emph{Follow-The-Leader} algorithm on strongly convex losses \citep{Kakade09} which picks the sequence of policies $\hat{\pi}_{1:N}$. Hence the presented results also hold for \emph{any} other no regret online learning algorithm we would apply to our imitation learning setting. In particular, we can consider the results here a reduction of imitation learning to no-regret online learning where we treat mini-batches of trajectories under a single policy as a single online-learning example. We first briefly review concepts of online learning and no regret that will be used for this analysis.
\subsection{Online Learning}
In online learning, an algorithm must provide a policy $\pi_n$ at iteration $n$ which incurs a loss $\ell_n(\pi_n)$. After observing this loss, the algorithm can provide a different policy $\pi_{n+1}$ for the next iteration which will incur loss $\ell_{n+1}(\pi_{n+1})$. The loss functions $\ell_{n+1}$ may vary in an unknown or even adversarial fashion over time. A no-regret algorithm is an algorithm that produces a sequence of policies $\pi_1, \pi_2, \dots, \pi_N$ such that the average regret with respect to the best policy in hindsight goes to 0 as $N$ goes to $\infty$:
\begin{equation}
\frac{1}{N} \sum_{i=1}^N \ell_i(\pi_i) - \min_{\pi \in \Pi} \frac{1}{N} \sum_{i=1}^N \ell_i(\pi) \leq \gamma_N
\end{equation}
for $\lim_{N \rightarrow \infty} \gamma_N = 0$. Many no-regret algorithms guarantee that $\gamma_N$ is $\tilde{O}(\frac{1}{N})$ (e.g. when $\ell$ is strongly convex) \citep{Hazan06, Kakade08, Kakade09}.
\subsection{No Regret Algorithms Guarantees}
Now we show that no-regret algorithms can be used to find a policy which has good performance guarantees under its own distribution of states in our imitation learning setting. To do so, we must choose the loss functions to be the loss under the distribution of states of the current policy chosen by the online algorithm: $\ell_i(\pi) = \mathbb{E}_{s \sim d_{\pi_i}}[\ell(s,\pi)]$.


For our analysis of \textsc{DAgger}, we need to bound the total variation distance between the distribution of states encountered by $\hat{\pi}_i$ and $\pi_i$,
which continues to call the expert. The following lemma is useful:
\begin{lem}
$||d_{\pi_i} - d_{\hat{\pi}_i}||_1 \leq 2 T \beta_i$.
\begin{proof}
Let $d$ the distribution of states over $T$ steps conditioned on $\pi_i$ picking $\pi^*$ at least once over $T$ steps. Since $\pi_i$ always executes $\hat{\pi}_i$ over $T$ steps with probability $(1-\beta_i)^T$ we have $d_{\pi_i} = (1-\beta_i)^T d_{\hat{\pi}_i} + (1-(1-\beta_i)^T) d$. Thus
\begin{displaymath}
\begin{array}{rl}
\multicolumn{2}{l}{||d_{\pi_i} - d_{\hat{\pi}_i}||_1}\\
= & (1-(1-\beta_i)^T) ||d - d_{\hat{\pi}_i}||_1\\
\leq & 2 (1-(1-\beta_i)^T)\\
\leq & 2 T \beta_i \\
\end{array}
\end{displaymath}
The last inequality follows from the fact that $(1-\beta)^T \geq 1 - \beta T$ for any $\beta \in [0,1]$.
\end{proof}
\end{lem}
This is only better than the trivial bound $||d_{\pi_i} - d_{\hat{\pi}_i}||_1 \leq 2$ for $\beta_i \leq \frac{1}{T}$. Assume $\beta_i$ is non-increasing and define $n_\beta$ the largest $n \leq N$ such that $\beta_n > \frac{1}{T}$. Let $\epsilon_N = \min_{\pi \in \Pi} \frac{1}{N} \sum_{i=1}^N \mathbb{E}_{s \sim d_{\pi_i}}[ \ell(s,\pi) ]$ the loss of the best policy in hindsight after $N$ iterations and let $\ell_{\max}$ be an upper bound on the loss, i.e. $\ell_i(s,\hat{\pi}_i) \leq \ell_{\max}$ for all policies $\hat{\pi}_i$, and state $s$ such that $d_{\hat{\pi}_i}(s) > 0$. We have the following:
\begin{theorem}\label{thmDagger2}
For \textsc{DAgger}, there exists a policy $\hat{\pi} \in \hat{\pi}_{1:N}$ s.t. $\mathbb{E}_{s \sim d_{\hat{\pi}}}[ \ell(s,\hat{\pi}) ] \leq \epsilon_N + \gamma_N + \frac{2 \ell_{\max}}{N} [n_\beta + T \sum_{i=n_\beta+1}^N \beta_i]$, for $\gamma_N$ the average regret of $\hat{\pi}_{1:N}$.
\begin{proof}
The last lemma implies $\mathbb{E}_{s \sim d_{\hat{\pi}_i}}(\ell_i(s,\hat{\pi}_i)) \leq \mathbb{E}_{s \sim d_{\pi_i}}(\ell_i(s,\hat{\pi}_i)) + 2 \ell_{\max} \min(1, T \beta_i)$. Then:
$\begin{array}{rl}
\multicolumn{2}{l}{\min_{\hat{\pi} \in \hat{\pi}_{1:N}} \mathbb{E}_{s \sim d_{\hat{\pi}}}[ \ell(s,\hat{\pi}) ]}\\
\leq & \frac{1}{N} \sum_{i=1}^N \mathbb{E}_{s \sim d_{\hat{\pi}_i}}(\ell(s,\hat{\pi}_i))\\
\leq & \frac{1}{N} \sum_{i=1}^N [ \mathbb{E}_{s \sim d_{\pi_i}}(\ell(s,\hat{\pi}_i)) + 2 \ell_{\max} \min(1, T \beta_i) ]\\
\leq & \gamma_N + \frac{2 \ell_{\max}}{N} [n_\beta + T \sum_{i=n_\beta+1}^N \beta_i] + \min_{\pi \in \Pi} \sum_{i=1}^N \ell_i(\pi)\\
= & \gamma_N + \epsilon_N + \frac{2 \ell_{\max}}{N} [n_\beta + T \sum_{i=n_\beta+1}^N \beta_i] \\
\end{array}$
\end{proof}
\end{theorem}
Under an error reduction assumption that for any input distribution, there is some policy $\pi \in \Pi$ that achieves surrogate loss of $\epsilon$, this implies we are guaranteed to find a policy $\hat{\pi}$ which achieves $\epsilon$ surrogate loss under its own state distribution in the limit, provided $\overline{\beta}_N \rightarrow 0$. For instance, if we choose $\beta_i$ to be of the form $(1-\alpha)^{i-1}$, then $\frac{1}{N}[n_\beta + T \sum_{i=n_\beta+1}^N \beta_i] \leq \frac{1}{N \alpha}[\log T + 1]$ and this extra penalty becomes negligible for $N$ as $\tilde{O}(T)$. As we need at least $\tilde{O}(T)$ iterations to make $\gamma_N$ negligible, the number of iterations required by \textsc{DAgger} is similar to that required by any no-regret algorithm. Note that this is not as strong as the general error or regret reductions considered in \citep{Beygelzimer05, Ross10, Daume09} which require only classification: we require a no-regret method or strongly convex surrogate loss function, a
stronger (albeit common) assumption. 

\paragraph{Finite Sample Case:} The previous results hold if the online learning algorithm observes the infinite sample loss, i.e. the loss on the true distribution of trajectories induced by the current policy $\pi_i$. In practice however the algorithm would only observe its loss on a small sample of trajectories at each iteration. We wish to bound the true loss under its own distribution of the best policy in the sequence as a function of the regret on the finite sample of trajectories.

At each iteration $i$, we assume the algorithm samples $m$ trajectories using $\pi_i$ and then observes the loss $\ell_i(\pi) = \mathbb{E}_{s \sim D_i}(\ell(s,\pi))$, for $D_i$ the dataset of those $m$ trajectories. The online learner guarantees $\frac{1}{N}\sum_{i=1}^N \mathbb{E}_{s \sim D_i}(\ell(s,\pi_i)) - \min_{\pi \in \Pi} \frac{1}{N}\sum_{i=1}^N \mathbb{E}_{s \sim D_i}(\ell(s,\pi)) \leq \gamma_N$. Let $\hat{\epsilon}_N = \min_{\pi \in \Pi} \frac{1}{N} \sum_{i=1}^N \mathbb{E}_{s \sim D_i}[ \ell(s,\pi) ]$ the training loss of the best policy in hindsight. Following a similar analysis to \citet{CesaBianchi04}, we obtain:
\begin{theorem}
For \textsc{DAgger}, with probability at least $1-\delta$, there exists a policy $\hat{\pi} \in \hat{\pi}_{1:N}$ s.t. $\mathbb{E}_{s \sim d_{\hat{\pi}}}[ \ell(s,\hat{\pi}) ] \leq \hat{\epsilon}_N + \gamma_N + \frac{2 \ell_{\max}}{N} [n_\beta + T \sum_{i=n_\beta+1}^N \beta_i] + \ell_{\max} \sqrt{\frac{2\log(1/\delta)}{mN}}$, for $\gamma_N$ the average regret of $\hat{\pi}_{1:N}$.
\begin{proof}
Let $Y_{ij}$ be the difference between the expected per step loss of $\hat{\pi}_i$ under state distribution $d_{\pi_i}$ and the average per step loss of $\hat{\pi}_i$ under the $j^{th}$ sample trajectory with $\pi_i$ at iteration $i$. The random variables $Y_{ij}$ over all $i \in \{1,2,\dots,N\}$ and $j \in \{1,2,\dots,m\}$ are all zero mean, bounded in $[-\ell_{\max}, \ell_{\max}]$ and form a martingale (considering the order $Y_{11}, Y_{12}, \dots, Y_{1m}, Y_{21}, \dots, Y_{Nm}$). By Azuma-Hoeffding's inequality $\frac{1}{mN} \sum_{i=1}^N \sum_{j=1}^m Y_{ij} \leq \ell_{\max} \sqrt{\frac{2\log(1/\delta)}{mN}}$ with probability at least $1-\delta$. Hence, we obtain that with probability at least $1-\delta$:
\begin{displaymath}
\begin{array}{rl}
\multicolumn{2}{l}{\min_{\hat{\pi} \in \hat{\pi}_{1:N}} \mathbb{E}_{s \sim d_{\hat{\pi}}}[ \ell(s,\hat{\pi}) ]}\\
\leq & \frac{1}{N} \sum_{i=1}^N \mathbb{E}_{s \sim d_{\hat{\pi}_i}}[ \ell(s,\hat{\pi}_i) ]\\
\leq & \frac{1}{N} \sum_{i=1}^N \mathbb{E}_{s \sim d_{\pi_i}}[ \ell(s,\hat{\pi}_i) ] + \frac{2 \ell_{\max}}{N} [n_\beta + T \sum_{i=n_\beta+1}^N \beta_i] \\
= & \frac{1}{N} \sum_{i=1}^N \mathbb{E}_{s \sim D_i}[ \ell(s,\hat{\pi}_i) ] + \frac{1}{mN}\sum_{i=1}^N \sum_{j=1}^m Y_{ij} \\
& + \frac{2 \ell_{\max}}{N} [n_\beta + T \sum_{i=n_\beta+1}^N \beta_i]\\
\leq & \frac{1}{N} \sum_{i=1}^N \mathbb{E}_{s \sim D_i}[ \ell(s,\hat{\pi}_i) ] + \ell_{\max} \sqrt{\frac{2\log(1/\delta)}{mN}} \\
& + \frac{2 \ell_{\max}}{N} [n_\beta + T \sum_{i=n_\beta+1}^N \beta_i]\\
\leq & \hat{\epsilon}_N + \gamma_N + \ell_{\max} \sqrt{\frac{2\log(1/\delta)}{mN}} + \frac{2 \ell_{\max}}{N} [n_\beta + T \sum_{i=n_\beta+1}^N \beta_i] \\
\end{array}
\end{displaymath}
\end{proof}
\end{theorem}
The use of Azuma-Hoeffding's inequality suggests we need $Nm$ in $O(T^2\log(1/\delta))$ for the generalization error to be $O(1/T)$ and negligible over $T$ steps. Leveraging the strong convexity of $\ell$ as in \citep{Kakade09} may lead to a tighter bound requiring only $O(T\log(T/\delta))$ trajectories.
%
%
\section{EXPERIMENTS} \label{sec:exp}
To demonstrate the efficacy and scalability of \textsc{DAgger}, we apply it to two challenging imitation learning problems and a sequence labeling task (handwriting recognition).
\subsection{Super Tux Kart}
\begin{figure}[htb]
\centering
\includegraphics[width=0.3\textwidth]{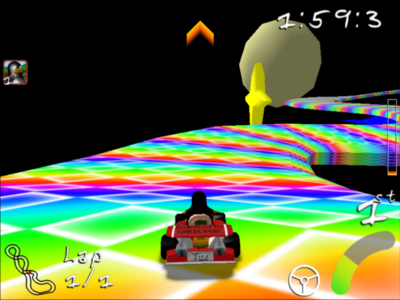}
\caption{Image from Super Tux Kart's Star Track.\label{figSTKScreenshot}}
\end{figure}

Super Tux Kart is a 3D racing game similar to the popular Mario Kart. Our goal is to train the computer to steer the kart moving at fixed speed on a particular race track, based on the current game image features as input (see Figure~\ref{figSTKScreenshot}). A human expert is used to provide demonstrations of the correct steering (analog joystick value in [-1,1]) for each of the observed game images.
For all methods, we use a linear controller as the base learner which updates the steering at 5Hz based on the vector of image features\footnote{Features $x$: LAB color values of each pixel in a 25x19 resized image of the 800x600 image; output steering: $\hat{y} = w^T x + b$ where $w$, $b$ minimizes ridge regression objective: $L(w,b) = \frac{1}{n} \sum_{i=1}^n (w^T x_i + b - y_i)^2 + \frac{\lambda}{2} w^T w$, for regularizer $\lambda=10^{-3}$.}.

\begin{figure}[htb]
\centering
\includegraphics[width=0.42\textwidth,trim=0 110 25 140,clip]{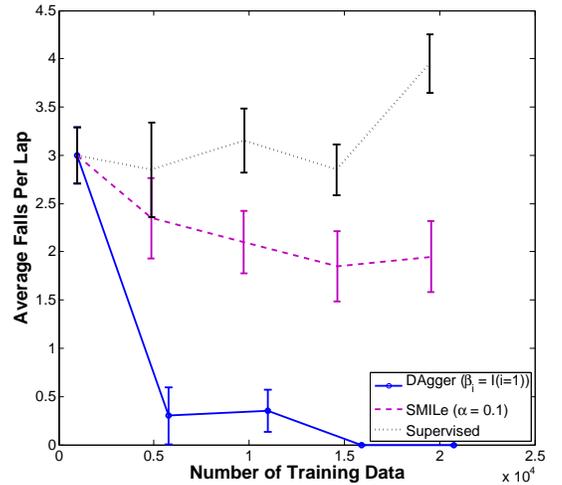}
\caption{Average falls/lap as a function of training data.\label{figResultSTK}}
\end{figure}

We compare performance on a race track called Star Track. As this track floats in space, the kart can fall off the track at any point (the kart is repositioned at the center of the track when this occurs). We measure performance in terms of the average number of falls per lap. For SMILe and \textsc{DAgger}, we used 1 lap of training per iteration ($\sim$1000 data points) and run both methods for 20 iterations. For SMILe we choose parameter $\alpha = 0.1$ as in \citet{Ross10}, and for \textsc{DAgger} the parameter $\beta_i = I(i=1)$ for $I$ the indicator function. Figure~\ref{figResultSTK} shows 95\% confidence intervals on the average falls per lap of each method after 1, 5, 10, 15 and 20 iterations as a function of the total number of training data collected.
We first observe that with the baseline supervised approach where training always occurs under the expert's trajectories that performance does not improve as more data is collected. This is because most of the training laps are all very similar and do not help the learner to learn how to recover from mistakes it makes. With SMILe we obtain some improvements but the policy after 20 iterations still falls off the track about twice per lap on average. This is in part due to the stochasticity of the policy which sometimes makes bad choices of actions. For \textsc{DAgger}, we were able to obtain a policy that never falls off the track after 15 iterations of training. Though even after 5 iterations, the policy we obtain almost never falls off the track and is significantly outperforming both SMILe and the baseline supervised approach. Furthermore, the policy obtained by \textsc{DAgger} is smoother and looks qualitatively better than the policy obtained with SMILe. A video available on YouTube \citep{Ross10a} shows a qualitative comparison of the behavior obtained with each method.
\subsection{Super Mario Bros.}
\begin{figure}[!htb]
\centering
\includegraphics[width=0.3\textwidth]{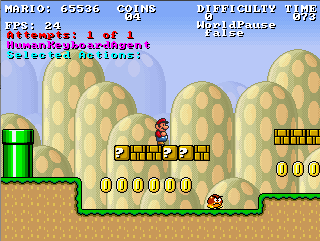}
\caption{Captured image from Super Mario Bros.\label{figMarioScreenshot}}
\end{figure}

Super Mario Bros. is a platform video game where the character, Mario, must move across each stage by avoiding being hit by enemies and falling into gaps, and before running out of time. We used the simulator from a recent Mario Bros. AI competition \citep{Togelius09} which can randomly generate stages of varying difficulty (more difficult gaps and types of enemies). Our goal is to train the computer to play this game based on the current game image features as input (see Figure~\ref{figMarioScreenshot}). Our expert in this scenario is a near-optimal planning algorithm that has full access to the game's internal state and can simulate exactly the consequence of future actions. An action consists of 4 binary variables indicating which subset of buttons we should press in $\{$left,right,jump,speed$\}$.
For all methods, we use 4 independent linear SVM as the base learner which update the 4 binary actions at 5Hz based on the vector of image features\footnote{For the input features $x$: each image is discretized in a grid of 22x22 cells centered around Mario; 14 binary features describe each cell (types of ground, enemies, blocks and other special items); a history of those features over the last 4 images is used, in addition to other features describing the last 6 actions and the state of Mario (small,big,fire,touches ground), for a total of 27152 binary features (very sparse). The $k^{th}$ output binary variable $\hat{y}_k = I(w_k^T x + b_k > 0)$, where $w_k,b_k$ optimizes the SVM objective with regularizer $\lambda = 10^{-4}$ using stochastic gradient descent \citep{Ratliff07, Bottou09}.}.

\begin{figure}[!htb]
\centering
\includegraphics[width=0.42\textwidth,trim=0 135 30 150,clip]{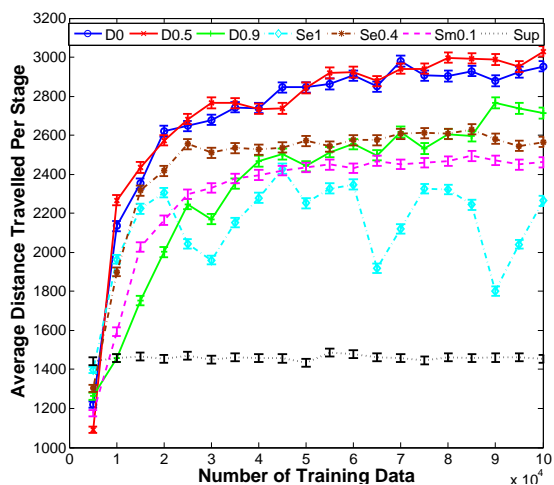}
\caption{Average distance/stage as a function of data.\label{figResultMario}}
\end{figure}

We compare performance in terms of the average distance travelled by Mario per stage before dying, running out of time or completing the stage, on randomly generated stages of difficulty 1 with a time limit of 60 seconds to complete the stage. The total distance of each stage varies but is around 4200-4300 on average, so performance can vary roughly in [0,4300]. Stages of difficulty 1 are fairly easy for an average human player but contain most types of enemies and gaps, except with fewer enemies and gaps than stages of harder difficulties. We compare performance of DAgger, SMILe and SEARN\footnote{We use the same cost-to-go approximation in \citet{Daume09}; in this case SMILe and SEARN differs only in how the weights in the mixture are updated at each iteration.} to the supervised approach (Sup). With each approach we collect 5000 data points per iteration (each stage is about 150 data points if run to completion) and run the methods for 20 iterations. For SMILe we choose parameter $\alpha = 0.1$ (Sm0.1) as in \citet{Ross10}. For \textsc{DAgger} we obtain results with different choice of the parameter $\beta_i$: 1) $\beta_i=I(i=1)$ for $I$ the indicator function (D0); 2) $\beta_i = p^{i-1}$ for all values of $p \in \{0.1,0.2,\dots,0.9\}$. We report the best results obtained with $p=0.5$ (D0.5). We also report the results with $p=0.9$ (D0.9) which shows the slower convergence of using the expert more frequently at later iterations. Similarly for SEARN, we obtain results with all choice of $\alpha$ in $\{0.1,0.2,\dots,1\}$. We report the best results obtained with $\alpha=0.4$ (Se0.4). We also report results with $\alpha = 1.0$ (Se1), which shows the unstability of such a pure policy iteration approach. Figure~\ref{figResultMario} shows 95\% confidence intervals on the average distance travelled per stage at each iteration as a function of the total number of training data collected.
Again here we observe that with the supervised approach, performance stagnates as we collect more data from the expert demonstrations, as this does not help the particular errors the learned controller makes. In particular, a reason the supervised approach gets such a low score is that under the learned controller, Mario is often stuck at some location against an obstacle instead of jumping over it. Since the expert always jumps over obstacles at a significant distance away, the controller did not learn how to get unstuck in situations where it is right next to an obstacle. On the other hand, all the other iterative methods perform much better as they eventually learn to get unstuck in those situations by encountering them at the later iterations. Again in this experiment, \textsc{DAgger} outperforms SMILe, and also outperforms SEARN for all choice of $\alpha$ we considered. When using $\beta_i=0.9^{i-1}$, convergence is significantly slower could have benefited from more iterations as performance was still improving at the end of the 20 iterations. Choosing $0.5^{i-1}$ yields slightly better performance (3030) then with the indicator function (2980). This is potentially due to the large number of data generated where mario is stuck at the same location in the early iterations when using the indicator; whereas using the expert a small fraction of the time still allows to observe those locations but also unstucks mario and makes it collect a wider variety of useful data. A video available on YouTube \citep{Ross10b} also shows a qualitative comparison of the behavior obtained with each method.
\subsection{Handwriting Recognition}
Finally, we demonstrate the efficacy of our approach on a structured prediction problem involving recognizing handwritten words given the sequence of images of each character in the word. We follow \citet{Daume09} in adopting a view of structured prediction as a degenerate form of imitation learning where the system dynamics are deterministic and trivial in simply passing on earlier predictions made as inputs for future predictions.
We use the dataset of \citet{Taskar03} which has been used extensively in the literature to compare several structured prediction approaches. This dataset contains roughly 6600 words (for a total of over 52000 characters) partitioned in 10 folds. We consider the large dataset experiment which consists of training on 9 folds and testing on 1 fold and repeating this over all folds. Performance is measured in terms of the character accuracy on the test folds.



\begin{figure}[!htb]
\centering
\includegraphics[width=0.42\textwidth,trim=0 110 30 125,clip]{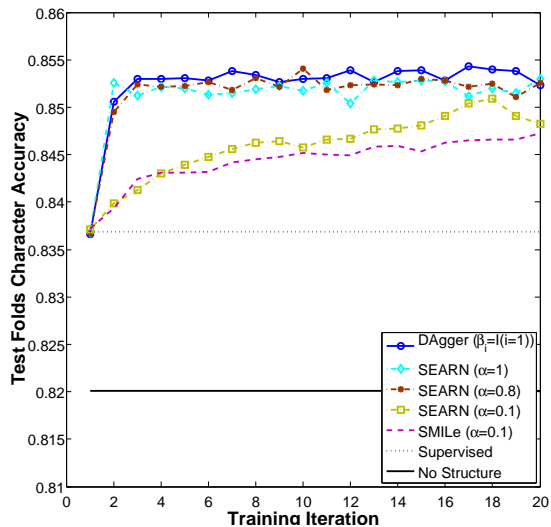}
\caption{Character accuracy as a function of iteration.\label{figResultOCR}}
\end{figure}

We consider predicting the word by predicting each character in sequence in a left to right order, using the previously predicted character to help predict the next and a linear SVM\footnote{Each character is 8x16 binary pixels (128 input features); 26 binary features are used to encode the previously predicted letter in the word. We train the multiclass SVM using the all-pairs reduction to binary classification \citep{Beygelzimer05}.}, following the greedy SEARN approach in \citet{Daume09}. Here we compare our method to SMILe, as well as SEARN (using the same approximations used in \citet{Daume09}). We also compare these approaches to two baseline, a non-structured approach which simply predicts each character independently and the supervised training approach where training is conducted with the previous character always correctly labeled. Again we try all choice of $\alpha \in \{0.1,0.2, \dots,1\}$ for SEARN, and report results for $\alpha = 0.1$, $\alpha=1$ (pure policy iteration) and the best $\alpha=0.8$, and run all approaches for 20 iterations. Figure \ref{figResultOCR} shows the performance of each approach on the test folds after each iteration as a function of training data.
The baseline result without structure achieves 82\% character accuracy by just using an SVM that predicts each character independently. When adding the previous character feature, but training with always the previous character correctly labeled (supervised approach), performance increases up to 83.6\%. Using DAgger increases performance further to 85.5\%. Surprisingly, we observe SEARN with $\alpha=1$, which is a pure policy iteration approach performs very well on this experiment, similarly to the best $\alpha=0.8$ and DAgger. Because there is only a small part of the input that is influenced by the current policy (the previous predicted character feature) this makes this approach not as unstable as in general reinforcement/imitation learning problems (as we saw in the previous experiment). SEARN and SMILe with small $\alpha=0.1$ performs similarly but significantly worse than DAgger. Note that we chose the simplest (greedy, one-pass) decoding to illustrate the benefits of the DAGGER approach with respect to existing reductions. Similar techniques can be applied to multi-pass or beam-search decoding leading to results that are competitive with the state-of-the-art.
%
%
%
\section{FUTURE WORK}
We show that by batching over iterations of interaction with a system, no-regret methods, including the presented \textsc{DAgger} approach can provide
a learning reduction with strong performance guarantees in both imitation learning and structured prediction. In future work, we will consider more sophisticated strategies
than simple greedy forward decoding for structured prediction, as well as using base classifiers that rely on Inverse Optimal Control \citep{Abbeel04, Ratliff06} techniques to learn a cost function for a planner to aid prediction in imitation learning.
%
%
Further we believe techniques similar to those presented, by leveraging a cost-to-go estimate, may provide an understanding of the success of online methods for reinforcement
learning and suggest a similar data-aggregation method that can guarantee performance in such settings.
%
%
%
%
\subsubsection*{Acknowledgements}
This work is supported by the ONR MURI grant N00014-09-1-1052, Reasoning in Reduced Information Spaces, and by the National Sciences and Engineering Research Council of Canada (NSERC).
\bibliographystyle{abbrvnat}
\bibliography{biblio}

\end{document}